\def\year{2017}\relax
\documentclass[letterpaper]{article} 
\usepackage{aaai17}  
\usepackage{times}  
\usepackage{helvet}  
\usepackage{courier}  
\usepackage{url}  
\usepackage{graphicx}  
\usepackage{listings}
\usepackage{algorithm}
\usepackage{amsmath}
\usepackage{xcolor}
\usepackage{subcaption}
\usepackage{float}
\usepackage{booktabs}
\definecolor{vkcolor}{RGB}{0,153,0}

\newcommand{\customlength}{\linewidth}
\definecolor{nucolor}{RGB}{153, 153, 0}

\begin{document}
%
\title{STARDATA: A StarCraft AI Research Dataset}
\author{
Zeming Lin\\
Facebook\\
770 Broadway\\
New York, NY, 10003
\And
Jonas Gehring\\
Facebook\\
6, rue M{\'e}nars\\
75002 Paris, France
\And
Vasil Khalidov\\
Facebook\\
6, rue M{\'e}nars\\
75002 Paris, France
\And
Gabriel Synnaeve\\
Facebook\\
770 Broadway\\
New York, NY, 10003
}
\maketitle
\begin{abstract}
We release a dataset of 65646 StarCraft replays that contains 1535 million frames and 496 million player actions. We provide full game state data along with the original replays that can be viewed in StarCraft. 
The game state data was recorded every 3 frames which ensures suitability for a wide variety of machine learning tasks such as strategy classification, inverse reinforcement learning, imitation learning, forward modeling, partial information extraction, and others.
We use TorchCraft to extract and store the data, which standardizes the data format for both reading from replays and reading directly from the game.
Furthermore, the data can be used on different operating systems and platforms.
The dataset contains valid, non-corrupted replays only and its quality and diversity was ensured by a number of heuristics.
We illustrate the diversity of the data with various statistics and provide examples of tasks that benefit from the dataset.
We make the dataset available at https://github.com/TorchCraft/StarData.
En Taro Adun!
\end{abstract}
\section{Introduction}

Real time strategy games as an AI research problem is attracting substantial attention~\cite{ontanon_survey_2013,usunier_episodic_2016,peng_multiagent_2017} due to their complex game dynamics, partial observability, and existing expert games in the form of human replays.
These games are a good test bed for various reinforcement learning algorithms on a domain with higher complexity than toy robotics tasks and turn-based board games.
Due to recent advances in deep learning, we see a trend of improved model performance with larger datasets.
As learning capacity of these models increases, there is a growing need for data, especially in order to apply deep learning methods to control in RTS games.


Although learning in StarCraft can be performed through playing, the dynamics of the game are extremely complex, and it is beneficial to speed up learning by using existing games.
The availability of datasets of recorded games between experienced players is therefore desirable.

StarCraft allows one to record replays of games which contain all commands issued by players. A number of online resources contain collections of replays from various tournaments (see Table~\ref{tab:related_work:datasets}).
Some information can be directly inferred from the replay file; however, reconstructing the full game state requires playback in StarCraft.

There are several aspects that make it difficult to use the replays directly for machine learning purposes. {\it Firstly}, the reconstruction speed of StarCraft is limited and would impose an upper threshold on training speed. {\it Secondly}, incompatibility between replays produced by different StarCraft versions makes it impossible to use the same game engine for all the replays or might result in corrupted data. {\it Finally}, the reconstruction process can only be reliably run on Windows, which adds additional unnecessary restrictions.
Hence, the utility of a replay dataset can be increased by extracting game states, validating them and storing them as a separate dataset.


For a dataset to serve as a good base for learning models, it should fulfill a number of requirements:
\begin{itemize}
\item[] {\bf Universality}: the data stored in the dataset can be used to learn different aspects of game strategy and at different levels. Thus the dataset should provide data which is not specific to any particular context and should be as close to the full game state as possible.
\item[] {\bf Diversity}: the dataset should cover a variety of game scenarios in terms of match-ups, maps, player strategies, etc.
\item[] {\bf Validity}: the dataset should be representative of the distribution of StarCraft matches where both sides are trying to win. 
\item[] {\bf Interfacing}: one should be able to easily substitute game states received from the game engine with game states recorded in the dataset.
\item[] {\bf Portability}: dataset access should be supported on a variety of platforms and operating systems.
\end{itemize}

With these requirements in mind, we constructed a new dataset of StarCraft replays from games among humans that can be used for StarCraft AI research. The following are our major contributions.

We provide a large set of StarCraft human replays, which is about 10x bigger than any of the comparable datasets currently available. The dataset includes a variety of scenarios and thus ensures the {\it diversity} requirement. Detailed statistics on matchups, maps etc. can be found in further sections.

All replays are checked for playability in StarCraft and BWAPI. We used additional scripted rule-based checks for corruption to fulfill the {\it validity} requirement.

The dataset is stored in a format that can be read by TorchCraft~\cite{synnaeve_torchcraft:_2016}, a library used as an interface between scientific computing frameworks and StarCraft. One can use exactly the same code to read data from the dataset and control StarCraft. This ensures both {\it interfacing} and {\it portability} requirements, since TorchCraft has a client in C++, Lua, and Python, and be compiled easily on any operating system.

\begin{table*}
\begin{center}
\begin{small}
\begin{tabular}{cccccc}
\toprule
Dataset                        & \# Replays & Source         &
Data Format & Extracted Data & Frame Skip \\
\midrule
\cite{weber_data_2009}         & 5493       & GG, TL, IC     &
text game logs & first production time &  n/a \\
\cite{synnaeve_dataset_2012}   & 7649       & GG, TL, IC     &
text game logs & full game data & 25 / 100 frames \\
\cite{cho_replay-based_2013}   & 5493+570   & GG, TL, IC, YG &
text game logs & first production time & n/a \\
\cite{robertson_improved_2014} & 7649       & GG, TL, IC     &
database & full game data & adaptive, 6 / 24 frames \\
STARDATA (this work)          & 65646      & GG, TL, IC, BR &
TorchCraft & full game data & 3 frames \\
\bottomrule
\end{tabular}
\end{small}
\\
\caption{\label{tab:related_work:datasets}Summary on StarCraft AI datasets. Sources: GG=\url{GosuGamers.net}, TL=\url{TeamLiquid.net}, IC=\url{ICCup.com}, YG=\url{ygosu.com}, BR=\url{bwreplays.com}}
\end{center}
\end{table*}

For each replay in the dataset, the complete game state is stored every 3 frames (about 8 frames per second). This means that one can employ the dataset to learn different aspects of the game strategy, from micro level to macro level and the {\it universality} requirement is fulfilled.

The current paper is structured as follows. First, we give an overview of the existing datasets, their main features and limitations. We further describe in detail our new dataset, how it was constructed and verified. Next we present some statistics related to this dataset and provide example scenarios for which it would be useful. We also provide metadata on where battles are in the dataset for unit micromanagement tasks. Discussion on further scenarios and use cases for this dataset concludes the paper.

\section{Related Work}
\label{sec:related_work}

The existing StarCraft datasets can be subdivided into two groups based on extracted data type (Table~\ref{tab:related_work:datasets}). Datasets from the first group, i.e.~\cite{hsieh_building_2008,weber_data_2009,cho_replay-based_2013} focus on specific aspects of the game and contain data that can only be used in a particular context. Datasets from the second group, e.g.~\cite{synnaeve_dataset_2012} and~\cite{robertson_improved_2014} contains general purpose full state data and is not restricted to any particular scope.

The datasets from the first group would typically target general strategy aspects. For example, the goal in~\cite{weber_data_2009} was to capture player's build orders, or strategic decisions timing. Thus for every replay it contains a log which specifies the first time each type of unit was produced by each player. Additionally, every replay was labeled by an expert, with labels that correspond to a predefined set of strategies. The replays in the dataset are taken from professional level tournaments.
\cite{cho_replay-based_2013} extended this set by including additional replays with unit visibility data. The datasets of this group tend to be compact, but their use is limited to build order prediction. 
Both~\cite{hsieh_building_2008},~\cite{kim_cooperative_2010}, and~\cite{synnaeve_bayesian_2011,synnaeve_opening_2011} have done build order prediction as a task.

The datasets from the second group would typically capture the full game state and tend to be more versatile. Their usability is determined by the completeness of state representation and frequency at which the states are saved. \cite{synnaeve_dataset_2012} contains game events, player commands, economical situation data for every 25 frames and unit location data for every 100 frames. Thus it is most suitable for medium and macro level AI modules. The replays in the dataset are also taken from professional level tournaments.
\cite{robertson_improved_2014} took the same set of replays and used an adaptive frame recording rate (every 24 or 6 frames) based on player actions. This made the dataset suitable for unit micromanagement tasks.

The summary on the existing datasets is given in Table~\ref{tab:related_work:datasets}. For each of the datasets we provide the number of replays included into the set, specify data sources, data format, type of data extracted and frequency at which the data is extracted.

Several authors have released works that would benefit from the use of a large scale and standardized game replay dataset.
\cite{hostetler_inferring_2012} manually collects a dataset of 509 games to do unit count prediction under partial observation in the early game.
\cite{uriarte_automatic_2015} uses the existing dataset provided by \cite{synnaeve_dataset_2012} to construct a combat model for micromanagement.
Recent advance in deep reinforcement learning could benefit from a large existing repository of high quality replays, as \cite{silver_mastering_2016} showed by using a large set of Go replays to create a system that could defeat the world champion of Go. \cite{usunier_episodic_2016},~\cite{peng_multiagent_2017}, and~\cite{foerster_stabilising_2017} all tackle the problem of micromanagement using deep reinforcement learning methods, which may be improved with this dataset of fine-grained game state data.

\section{Dataset}
\label{sec:dataset}
In what follows  we refer to the StarCraft recorded games as the \emph{original replays},
and the TorchCraft recorded game states as \emph{extracted replays}. 

The new STARDATA dataset contains 65546 original and extracted replays, almost 10 times bigger than the largest existing datasets (see Table~\ref{tab:related_work:datasets}). 
It is based on replays from~\cite{synnaeve_dataset_2012} and uses an additional source of replays (BR, \url{bwreplays.com})). 
There is no restriction on replay submission on this site, so some replays may be corrupt, i.e. they might not correspond to a typical game scenario where both sides are trying to win.
There are two cases when corruption occurs: (1) the replay is recorded on an older version of the game engine, and (2) the replay is produced by players testing out strategies and game dynamics instead of trying to win.

The original replays only store a list of commands issued by each player. If they are played through a different version of StarCraft, the replay may become corrupt, resulting in a game state that is essentially halted or looping until one player leaves the game. 
This happens because, for example, a player loses a battle due to a change in unit strengths in a different patch and begins trying to control a non-existent army.
The other source of corruption occurs when players play for any other reason except to win the game. For example two players could be testing out strategies and decide to cooperatively achieve a specific game state.

To ensure game quality, we (1) remove games produced before November 25, 2008 -- the release date for StarCraft patch 1.16, and (2) implemented a heuristic that would track mineral and gas usage in the second half of the game and filter out those that are under 70\% of total collected amounts.
Games may be played on older patches after the date patch 1.16 is release.
On visual inspection of several games like these, we saw that the heuristic would filter out most of them.
The false positives of this heuristic tend to be highly unbalanced games, or games with players who are not familiar with the rules of StarCraft and thus cannot manage their resources well. 

The saved frame data contains full game state, including unit dynamic characteristics (about 30 per unit) and bullets. Every third frame is saved, making the dataset suitable for tasks that require high frequency control, such as micro management. However, we encode frames as their deltas (differences), and we apply the ZStandard (lossless) compression algorithm. This gives us a 38x compression ratio over storing the full state every 3 frames, while enabling lossless reconstruction. The STARDATA set total size is 365 GB (compressed, can be streamed), while the original replays' total size is 5.5 GB. To provide a common evaluation ground, we split the dataset randomly into subsets: \emph{train} (59060 games), \emph{development} (3289 games) and \emph{test} (3297 games).

\section{Statistics}
\label{sec:statistics}

\begin{table}
\begin{center}
\begin{tabular}{cccccc}
\toprule
PvZ & TvZ & PvT & TvT & PvP & ZvZ \\
18016  & 14531  & 17385  & 2550   & 7015   & 6149   \\
\bottomrule
\end{tabular}
\\
\caption{\label{tab:statistics:matchups}Number of games per matchup in STARDATA.
Legend: P = Protoss, T = Terran, Z = Zerg}
\end{center}
\end{table}

\begin{table}
\begin{center}
\begin{tabular}{cc|cc}
\toprule
Map Name             & Count & Map Name         & Count \\
\midrule
Fighting Spirit      & 19817 & Othello 1.1      & 1161 \\
Python 1.3           & 7545  & Colosseum II 2.0 & 1086 \\
Heartbreak Ridge     & 5700  & Longinus         & 953  \\
Destination 1.1      & 5655  & Andromeda 1.0    & 808  \\
Tau Cross            & 1383  & Icarus 1.0       & 769  \\
Circuit Breakers 1.0 & 1366  & Medusa 2.2       & 763  \\
Blue Storm           & 1269  & Outsider 2.0     & 716  \\
\bottomrule
\end{tabular}
\caption{\label{tab:statistics:maps}Most frequent maps in STARDATA.}
\end{center}
\end{table}

Scenario diversity is summarized in Tables~\ref{tab:statistics:matchups} and~\ref{tab:statistics:maps} where we provide the distribution of games over matchups and map occurrences respectively. The most popular matchups are PvZ and PvT, and the least popular is TvT. 
Mirror matches are played far less often than matches with different races.
The most popular map is Fighting Spirit by far, followed by Python 1.3. The dataset includes
524 unique maps.
Two maps are considered the same if every tile has the same ground height and walkability.
182 maps were played more than 10 times per map,
and 83 maps were played more than 100 times per map. The tail is composed of older versions of maps and a few ones rarely seen in competitive play. We decided to include them anyway for the sake of diversity.
We do not guarantee that all maps in the dev and test sets appear in the train set. We believe it is important to be able to understand a map based on the statistics of other similar maps. Although humans will have trouble playing a new map from scratch, they are able to quickly pick up intricacies and succeed on them to some degree.
Since most of our dataset comes from a public database, we cannot provide any player disambiguation statistics, since players are free to change their username as much as they want in StarCraft.

\subsection{Game statistics}

\begin{figure}
\centering
\includegraphics[width=\customlength]{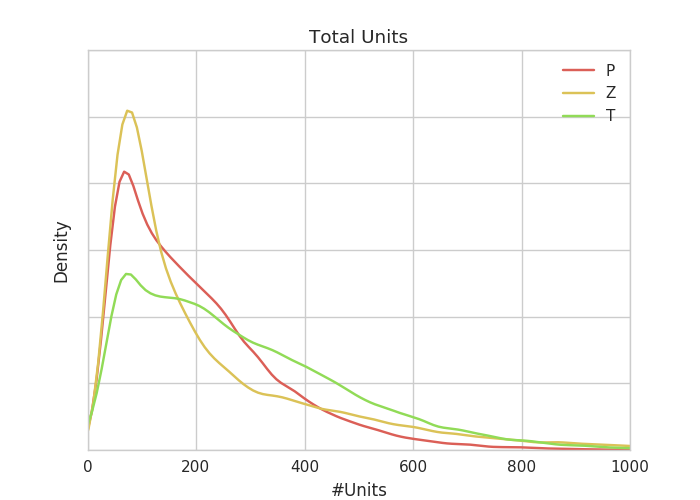}
\caption{
 \label{fig:statistics:units}
 Density plot of the total number of units created.  A few outliers of more than 1000 units are not shown. 
}
\end{figure}

\begin{figure}
\centering
\includegraphics[width=\customlength]{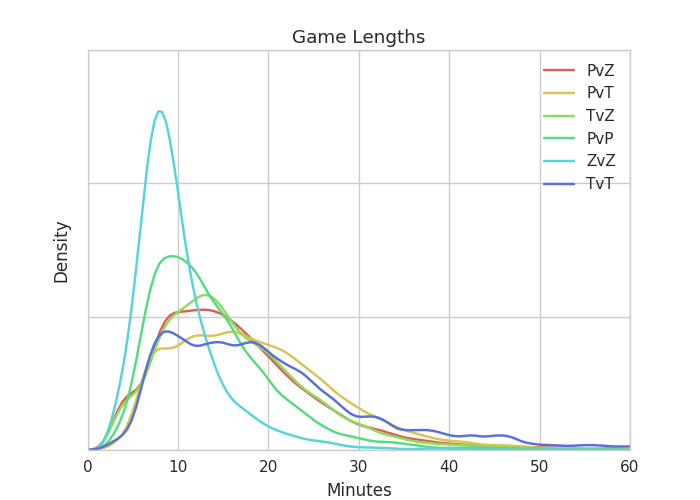}
\caption{
 \label{fig:statistics:lengths}
 Density plot of game lengths in minutes. A few outliers that last longer than 60 minutes are not shown.
 }
\end{figure}

Figures~\ref{fig:statistics:units} and~\ref{fig:statistics:lengths} show density plots of the total number of units produced and game lengths respectively.
Games seem to follow a log-normal distribution shape, with most games lasting between 10 and 20 minutes and producing between 40 to 300 units.
There is a long tail of games over 60 minutes and 1000 units that we do not show.
We make a few interesting observations from these density plots:
(1) Most ZvZ matchups tend to finish early.
(2) TvT and PvT games generally last very long.
(3) Zerg games tend to end with few units or with many. We observe a very skewed distribution towards 0, but also a very fat tail.
(4) Protoss builds the fewest number of units, especially in the late game

\begin{figure}[t]
\centering
\includegraphics[width=\customlength]{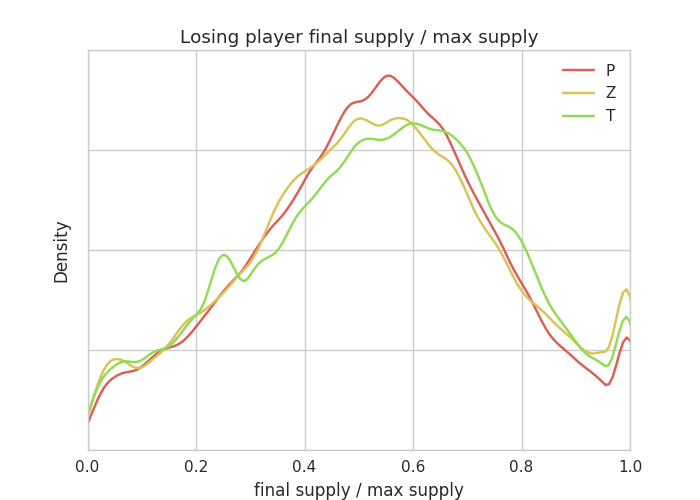}
\caption{
  \label{fig:statistics:units_psi}
  Ratio of effective supply to maximum supply of a surrendering player. A number near 0 indicates that the player fought until the bitter end. A number near 1 indicates that the player gave up fairly fast.
}
\end{figure}
 
Figure \ref{fig:statistics:units_psi} illustrates how easily games are given up.
It assumes the player with less supply at the end of the game is losing, which is generally true.
We find that the distribution of this ratio is bell-shaped and approximately normal. There is a spike at ratio 1.0 which corresponds to players who give up at their maximum supply. It is the most prominent for Zerg players and least for Protoss players.


\begin{figure*}
  \begin{subfigure}{0.5\textwidth}
  \centering
  \includegraphics[width=\customlength]{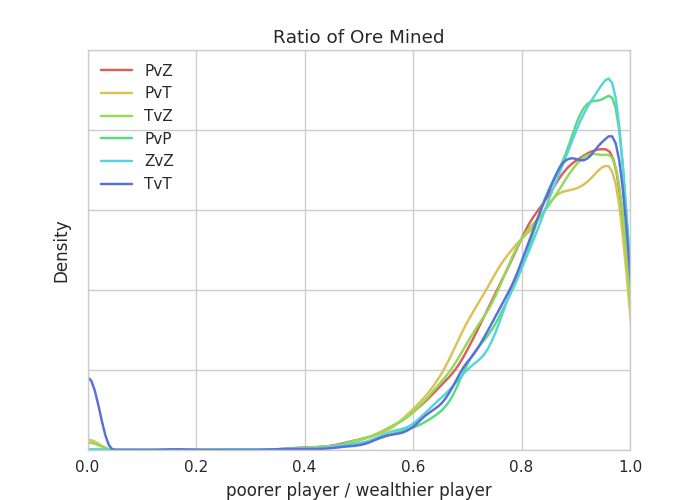}
  \end{subfigure} \hfill
  \begin{subfigure}{0.5\textwidth}
  \centering
  \includegraphics[width=\customlength]{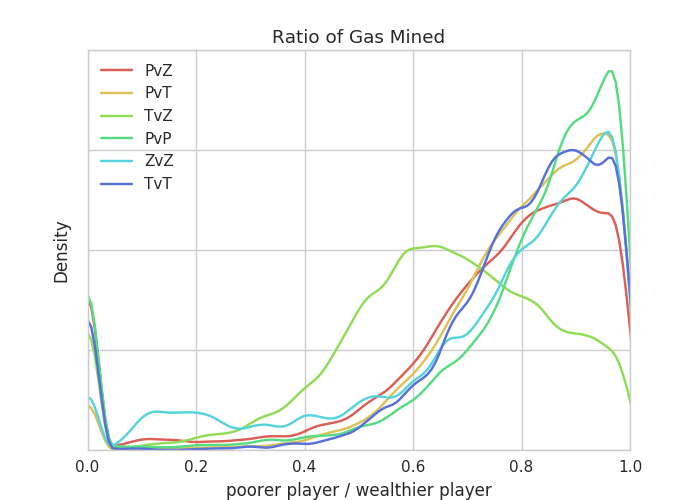}
  \end{subfigure}
\caption{
  \label{fig:statistics:ratio_res}
  Density plot of the ratio of resources gathered by the player with the most resources to the player with the least resources. A number closer to 1 indicates a more balanced game. "Rush" strategies typically have poor resource mining, which creates an apparent spike near 0.
}
\end{figure*}

Figure~\ref{fig:statistics:ratio_res} shows the ratio of resources mined between the two players (poorer to wealthier). This graph gives a good idea of how balanced the
dataset is in terms of skill level. If players mine approximately the same number of resources as each other, then their skills are probably similar. We see a distribution much like what
we would expect, with most games close to 1. We see that the most unbalanced games tend to be
TvT, PvT, or TvZ. Several games of these three matchups see one side mining 20x the resources
of the other side. This is likely due to how strong a Terran defense can be, as they defend
against attacks on relatively few resources and with no map control. 
We see that a significant proportion of all games ends with no gas mined, as they end too early due  to a "rush" strategy. 
We also observe that most TvZ games end with one player mining twice as much gas than the other, likely due to Zerg's high gas dependence.

\section{Downstream Tasks}

We provide some examples of learning and control tasks that can be addressed with the use of this dataset.

\subsection{Strategy Classification}
The simplest form of strategy classification is to predict build order, i.e. order in which buildings are constructed. And the major challenge is to perform the prediction under partial observations.
We provided many examples of existing research into this domain in Related Works.
Since this dataset contains full game states and provides a large diversity in terms of strategies employed, it should serve as a good based for strategy space estimation and strategy inference.

\subsection{Inverse Reinforcement Learning}
Inverse reinforcement learning (IRL), or apprenticeship learning~\cite{abbeel_apprenticeship_2004}, is the process of learning a task when given an expert trace but no explicit reward function.
Although StarCraft does have a reward function as 0--1 win--loss score at the end of a game, the rewards are quite sparse and learning a more accurate reward function may be necessary for control.
Furthermore, this dataset provides a large repository of auxiliary signals, such as number of units, resources, and more abstract ideas like map control to help ease the problem.
However, doing IRL in an environment with a huge action space and state space under partial observability is still a difficult task.

\subsection{Imitation learning}
Imitation learning is the process of learning to perform a task given only a demonstration of the task.
Given a few state sequences, the goal is to find a mapping function $f() : Z \rightarrow A$ that returns an action for each state and demonstrations the behavior shown.
Some key problems include dealing with unseen states, generalizing to new situations, and poor data quality \cite{argall_survey_2009}
As for IRL, this dataset provides a rich dataset with complex dynamics as a test bed for imitation learning algorithms, without the expense involved with real robots.

\subsection{Forward modeling}
Forward modeling is essentially predicting the future.
A recent direction focuses on predicting future video frames from past frames \cite{mathieu_deep_2015}.
This dataset provides a level of complexity between turn based board games and the real world, as video prediction cannot rely on a structured distribution of future frames as StarCraft games can.
Having a good forward model would allow a StarCraft AI to make tactical decisions, and forward modeling research benefits from the restricted domain and complex dynamics that StarCraft provides.
Additionally, a parameterized forward model may allow us to skip an expensive simulator for a single step of a forward model, or step forward in time cheaply to enable better tree search on large action and state spaces.
Since this dataset provides full game states at high sampling rates, it is suitable for forward modeling tasks at different scales. One can learn a forward model on very small time steps, to see if a model can learn the short-term dynamics of a game.
\cite{uriarte_automatic_2015} used a replay dataset to learn a combat model of StarCraft dynamics.
One can also learn a forward model operating on the long-term dynamics, predicting the tactics and unit distributions across the game map.

\subsection{Partial Information Handling}
Another difficult problem is to derive and adjust player's strategy under partial observability in StarCraft. Example tasks include: (1) uncover the fog of war and restore the complete state using partial observations; (2) derive models to control observability; (3) exploit partial observability to hide your strategy.
For example, \cite{weber2011particle} uses particle filters, and \cite{synnaeve_bayesian_2012} uses Kalman filters, both to estimate the position of enemy units under the fog of war after seeing them for briefly.
Professional human players are very efficient in handling partial observations and solving the mentioned tasks. And the capacity to handle partial information is an important constituent for the successful strategy in StarCraft. This dataset should serve as a good base for approaching such tasks.

\section{Battle detection}
\begin{algorithm}[ht]
\begin{enumerate}
\item Find the location of every death in the game
\item Perform Mean Shift with flat kernel of size 1 on the deaths
    in \texttt{x-y-t} space, normalizing such that a unit ball corresponds
    to 20 seconds on \texttt{t} axis and 200 walktiles on \texttt{x} and
    \texttt{y} axes.
\item Merge modes that are within a unit ball from each other, assign unit
    death events to the closest modes within a unit ball.
\item Filter out all clusters with fewer than 3 deaths.
\item Add 6 seconds before and after each cluster on the \texttt{t} axis
    to be sure to include tactical maneuvers before and after the battle
\item Set battle bounding box to a 200x200 rectangle around the cluster center
    in \texttt{x-y} space.
\item Greedily merge all clusters with bounding boxes in \texttt{x-y-t} space 
	whose Jaccard similarity scores are greater than 0.6, restricting the \texttt{x-y} space
    to be a 200x200 rectangle
\end{enumerate}
\caption{\label{alg:analysis:battle_detection:battle_detection}Battle detection}
\end{algorithm}

\label{sec:battle_detection}
Unit micromanagement plays an important role in StarCraft. However, finding relevant battle segments in replays is a tedious process. \cite{synnaeve_dataset_2012} used a heuristic to backtrack the battles starting from the deaths of units, seeking agglomerative locality in space and time around the event, but was dependent on several hyperparameters. 
Thus we developed a procedure to detect the battles automatically and provide this information along with the dataset. We use Mean-Shift \cite{comaniciu2002mean} to cluster the deaths into several bins, and run some heuristic filters on the results. The full detection algorithm is described in algorithm \ref{alg:analysis:battle_detection:battle_detection}

We distribute this metadata: for each battle we specify battle location as a bounding box of the battle region, and duration as a time interval between the start frame and the end frame. We also provide the counts of each unit type involved in the battle.

\section{Opening clustering}
\begin{figure}[H]
\centering
\includegraphics[width=0.8\columnwidth]{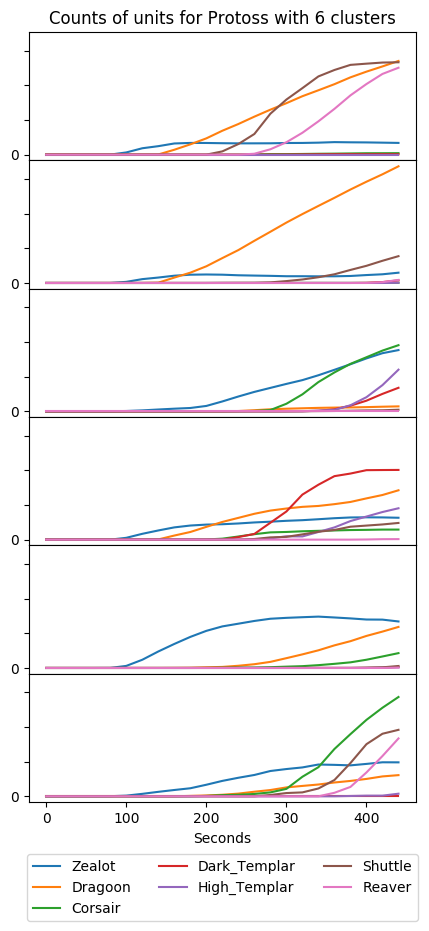}
\caption{
 \label{fig:op_cluster_plots}
 Example of the clusters obtained from numbers of each unit type, sampled every 20s, for Protoss. All units were used in the clustering, but we only display a subset for visual analysis. Each subplot is its own cluster. The y-axis can be interpreted as the normalized score of how often each unit is created in each time bin across all games. At the same time bin for the same unit type, a value twice as high implies the cluster mean had twice the number of units for that type.
}
\end{figure}

As an example of what exploratory data analysis can yield, inspired from \cite{synnaeve_opening_2011}, we performed clustering to in search for canonical opening strategies. 
We applied K-means over the flattened matrix of the number of units of each type, sampled every 20 seconds, for the first 8 minutes of each game.
We normalize the unit counts across each time bin across each type.
We cross-validated the number of clusters to maximize the silhouette coefficient, that measures the separation distance (intra-inter) of the clusters. Figure~\ref{fig:op_cluster_plots} exhibits some of the clusters we obtained for the Protoss race.

We can clearly observe some common Protoss strategies discovered via this analysis. 
The top most cluster appears to be a Dragoon heavy start with Reavers and Shuttles, a common strategy in PvP games. 
The third cluster is a Zealot heavy army with Cosair and High Templar support, a common strategy against Mutalisk heavy Zerg lineups.
The 4th cluster is a Dark Templar rush, a devastating strategy when the opponent does not counter them quickly enough with detection. 

\section{Conclusion}
\label{sec:conclusion}

We presented STARDATA -- a large dataset of StarCraft replays that contains full state data recorded every 3 frames. This enables the study of tactical and strategic elements of StarCraft gameplay at different scales. None of the existing datasets offers full state data on such a fine grain scale. One can address unit micromanagement scenarios that require high frequency control and macroeconomic strategy learning at the same time using this dataset.

STARDATA is 10x larger than the largest of the existing StarCraft datasets. It contains diverse game scenarios in terms of maps, matchups and player strategies -- we presented various statistics over the set to illustrate the diversity.
At the same time, we used several heuristics to keep only valid scenarios and filter out the corrupted ones.
The TorchCraft library was used to extract and store the data. This way we provide a standardized interface for access and ensure portability across platforms.

We believe that this dataset is going to be useful for the AI research community and propose a number of tasks for which this dataset can be employed. In particular, data-hungry deep reinforcement learning algorithms could benefit from a large amount of diverse scenarios. We also encourage others to propose interesting tasks to use this dataset as a benchmark.

\nocite{noauthor_bwreplays.com_2012}

\bibliographystyle{aaai}
\bibliography{stardata.bib}

\end{document}